\documentclass[preprint,12pt]{elsarticle}
 
\usepackage{lineno,hyperref}
\usepackage{amssymb}
\usepackage{graphicx}
\usepackage{amsmath}
\usepackage{subcaption}
\usepackage{color}
\usepackage{amsfonts}
\usepackage{float}
\usepackage{multirow}
\usepackage[linesnumbered,ruled,vlined]{algorithm2e}
\usepackage{textcomp}
\usepackage{times}
\usepackage{rotating,tabularx,siunitx}
\usepackage{slashbox}
\usepackage{makecell}

\modulolinenumbers[5]

\journal{}

\begin{document}

\begin{frontmatter}

\title{A Face Recognition Signature Combining Patch-based Features with Soft Facial Attributes}


\author[]{L.~Zhang}
\ead{lzhang@34@uh.edu}
\author[]{P.~Dou}
\ead{pdou@central.uh.edu}
\author[]{I.A.~Kakadiaris}
\ead{ioannisk@uh.edu}
 
\address{Computational Biomedicine Lab, 4849 Calhoun Rd, Rm 373, Houston, TX 77204}

\begin{abstract}
This paper focuses on improving face recognition performance with a new signature combining implicit facial features with explicit soft facial attributes. This signature has two components: the existing patch-based features and the soft facial attributes. A deep convolutional neural network adapted from state-of-the-art networks is used to learn the soft facial attributes. Then, a signature matcher is introduced that merges the contributions of both patch-based features and the facial attributes. In this matcher, the matching scores computed from patch-based features and the facial attributes are combined to obtain a final matching score. The matcher is also extended so that different weights are assigned to different facial attributes. The proposed signature and matcher have been evaluated with the UR2D system on the UHDB31 and IJB-A datasets. The experimental results indicate that the proposed signature achieve better performance than using only patch-based features. The Rank-1 accuracy is improved significantly by 4\% and 0.37\% on the two datasets when compared with the UR2D system.
\end{abstract}


\end{frontmatter}


\section{Introduction}

Face recognition is one of the major visual recognition tasks in the fields of biometrics, computer vision, image processing and machine learning. In recent years, most of the significant advances in visual recognition have been achieved by deep learning models, especially deep Convolutional Neural Networks (CNNs) \cite{ILSVRC2015, girshick2016region, szegedy2013deep}. CNN was first proposed in the late 1990s by LeCun \textit{et al.} \cite{lecun1989backpropagation, lecunhandwritten}. It was quickly overwhelmed by the combination of other shallow descriptors (such as SIFT, HOG, bag of words) with Support Vector Machines (SVMs). With the increase of image recognition data size and computation power, CNN has become more and more popular and dominant in the last five years. Krizhevsky \textit{et al.} \cite{Alex2012} proposed the classic eight-layer CNN model (AlexNet) with five convolutional and three fully connected layers. The model is trained via back-propagation through layers and performs extremely well in domains with a large amount of training data. Since then, many new CNN models have been constructed with larger sizes and different architectures to improve performance. Simonyan \textit{et al.} \cite{simonyan2014very} explored the influence of CNN depth by an architecture with small convolutional filters ($3 \times 3$). They achieved a significant improvement by pushing the depth to 16-19 layers in a VGG model. Szegedy \textit{et al.} \cite{szegedy2015going} introduced GoogLeNet as a 22-layer Inception network, which achieved impressive results in both image classification and object detection tasks. He \textit{et al.} \cite{He_2016_CVPR} proposed Residual Networks (ResNet) with a depth of up to 152 layers, which set new records for many image recognition tasks. Furthermore, He \textit{et al.} \cite{he2016identity2} proposed a residual network of 1,000 layers with identity mappings that makes training easier and improves generalization.
 
Recently, many CNNs have been introduced in face recognition and have achieved a series of breakthroughs. Similar to image recognition, effective CNNs require a larger amount of training images and larger network sizes. Yaniv \textit{et al.} \cite{taigman2014deepface} trained the DeepFace system with a standard eight-layer CNN using 4.4M labeled face images. Sun \textit{et al.}  \cite{sun2014deep, sun2014deep2, sun2015deepid3} proposed the Deep-ID systems with more elaborate network architectures and fewer training face images, which achieved better performance when compared with the DeepFace system. FaceNet \cite{schroff2015facenet} was introduced with 22 layers based on the Inception network \cite{szegedy2015going, zeiler2014visualizing}. It was trained on 200M face images and achieved further improvement. Parkhi \textit{et al.} \cite{parkhi2015deep} introduced the VGG-Face network with up to 19 layers adapted from Simonyan \textit{et al.} \cite{simonyan2014very}, which was trained on 2.6M images. This network also achieved comparable results and has been extended to other applications. To overcome the massive request of labeled training data, Masi \textit{et al.} \cite{masi16dowe} proposed to use domain specific data augmentation, which generates synthesis images for CASIA WebFace collection \cite{yi2014learning} based on different facial appearance variations. Their results trained with ResNet match the state-of-the-art results reported by the networks trained on millions of images. Recently, Xu \textit{et al.} \cite{xiang2017ijcb} presented the evaluation of a pose-invariant 3D-aided 2D face recognition system (UR2D) that is robust to pose variations as large as 90\textdegree{}. Different CNNs are integrated in face detection, landmark detection, 3D reconstruction and signature generation. 

Overall, previous deep face networks have two limitations: (a) the learned deep features is implicit; there is no human-readable information and the discriminative information is encoded in high dimensional features. (b) explicit soft facial attribute features are underestimated, which can be used to improve recognition performance. 
   
   \begin{figure} 
     \centering
   \begin{center}
     \includegraphics[width=1\linewidth]{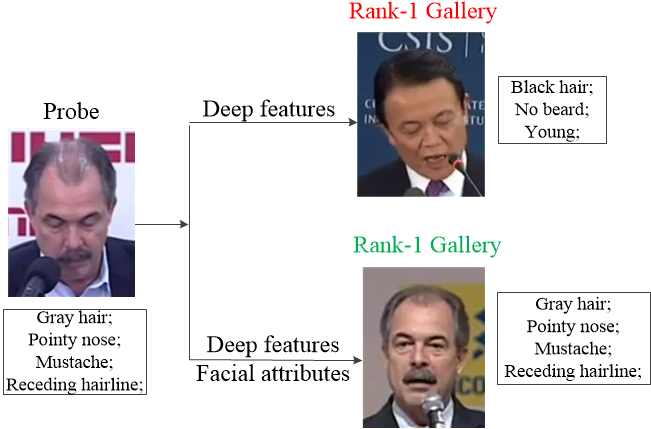}
   \end{center}
      \caption{Depicted a matching comparison based on implicit deep features and the combination of both implicit deep features and explicit soft facial attributes on the UR2D system. The upper gallery indicates the matching with only implicit deep features while the bottom gallery indicates the matching with both implicit deep features and explicit soft facial attributes.}
   \label{f1}
   \end{figure}
   
This work overcomes both limitations and improves face recognition performance by combining the implicit facial features with the new explicit facial attribute features. Due to the uncontrolled environments, the existing facial features are sensitive and more likely be corrupted. Thus, only using implicit facial features may lead to incorrect matching. On the other hand, the soft facial attribute features related to hair, eyebrow, nose and mouth are less sensitive to the change of environments and occlusions. By combining both facial features and soft facial attributes, more robust matching result can be obtained. Using the IJB-A face recognition dataset \cite{klare2015pushing} as an example, Figure~\ref{f1} depicts the intuition of the proposed method. It can be observed that, due to similar background and poses, the probe face is mis-matched based on implicit deep features in the upper matching gallery. However, the probe face does not share any facial attributes with the incorrect matching gallery image. Using the proposed signature, however, both implicit deep features and explicit facial attribute features are taken into account, which leads to a correct matching gallery image in the bottom matching result. It can be observed that these human-describable facial attributes can be used to fix the shortage of deep feature based face recognition system.

The contributions of this paper are improving face recognition performance and robustness by the following new signature and matcher: (i) a facial signature with two components of patch-based features and soft facial attributes. The facial attribute component is extracted by a CNN adapted from state-of-art networks. The contribution of facial attribute information is not well explored in deep feature based face recognition systems. (ii) a signature matcher combining the soft facial attribute component with the existing patch-based features in the UR2D system. The signature matcher is also extended to assign different weights to different facial attributes.    

The rest of this paper is organized as follows: Section \ref{sec2} presents the related work. Section \ref{sec3} and Section \ref{sec4} describe the proposed signature and matcher, respectively. The experimental design, results, and analysis are presented in Section \ref{sec5}. Section \ref{sec6} concludes the paper. 

\section{Related Work}
\label{sec2}

In face recognition, both global and local feature based methods have been proposed. Global methods learn discriminative information from the whole face image, such as subspace methods \cite{turk1991eigenfaces, belhumeur1997eigenfaces}, Sparse Representation based Classification (SRC) \cite{wright2009robust,yang2011robust} and Collaborative Representation based Classification (CRC) \cite{zhu2012multi,zhang2011sparse} and CNN based methods \cite{parkhi2015deep, masi16dowe}. Although global methods have achieved great success in controlled environments, they are sensitive to the variations of facial expression, illumination and occlusion in uncontrolled real-world scenarios. On the other hand, local methods extract features from local regions. The classic local features include Local Binary Patterns (LBP) \cite{ahonen2006face, liao2007learning}, Gabor features \cite{zhang2005local, su2009hierarchical}, Scale-Invariant Feature Transform (SIFT) \cite{luo2007person, bicego2006use}, gray values and so on. In local methods, most efforts focus on patch (block) based methods, which usually involve steps of local patch partition, local feature extraction, and local matching combination. With intelligent combination, these methods weaken the influence of variant-prone or occluded patches and ensemble the matching of invariant or unoccluded patches \cite{martinez2002recognizing, su2009hierarchical, yuk2011multi, zhang2015icb}. The drawback of most patch based methods is that they still rely on implicit features. The discriminative information is embedded in human-unreadable features. 

Soft Attributes, also known as high-level semantic features, have drawn a lot of attention in the past years in domains from image recognition to face recognition. In the face recognition domain, soft facial attributes like gender, race, age, hair color, and facial hair are very intuitive and they provide more human-understandable descriptions of a subject. Humans also rely on these intuitive attributes to remember and identify different persons. 

Many methods have been developed for improving the accuracy of attribute prediction. Based on the success of deep learning, most of them rely on CNNs. Yi \textit{et al.} \cite{yi2014age} proposed a multi-scale network that uses multiple local image patches as input to estimate age, gender, and ethnicity. The multi-scale image patches are cropped based on landmark locations. Liu \textit{et al.} \cite{liu2015faceattributes} developed the first CNN based facial attribute prediction framework to estimate 40 facial attributes. LNet and Anet are cascaded for face localization and attribute predication, respectively. Kang \textit{et al.} \cite{kang2015face} developed a face attribute classification method based on attribute-aware correlation maps and gated CNNs. Each correlation map of an attribute provides information about regions where the relevant features should be extracted. The CNN trained for each region is gated so that the classification errors of less relevant attributes contribute less in the learning process of back propagation. Rudd \textit{et al.} \cite{rudd2016moon} proposed a mixed objective optimization network based on joint optimization over all the attributes. The tasks of multi-label classification and domain adaptation are optimized under one unified objective function. Zhong \textit{et al.} \cite{zhong2016face} proposed to use off-the-shelf CNN architectures to extract features for attribute prediction. Considering the diversity of different attributes, the features are extracted from different levels in CNNs. Kalayeh  \textit{et al.} \cite{kalayeh2017improving} developed a semantic segmentation based network for facial attribute prediction. The localization cues learned by the semantic segmentation are used to guide the attention of the attribute prediction. Overall, these methods all focus on improving the prediction of facial attributes as a multi-label classification problem. In this paper, a further step is taken, which is using deep facial attributes to improve face recognition performance. Extracting high level semantic features from face recognition has been a goal for a long time. The facial attributes are both human-readable and describable. However, only using soft facial attributes may neglect the non-describable features, which can be captured effectively by CNNs. Therefore, in this paper, a signature is proposed to combine both patch-based features and facial attribute features to boost face recognition performance.  

Multi-label classification \cite{ji2010shared, zhu2016block, wang2016cnn,zhang2014fully, Zhang201789} is also related to the proposed method. In these methods, a classification model is learned to estimate multiple labels. These labels are used as output directly. On the other hand in the proposed method, the multi-labels are applied as input features to learn the matching identity.

\section{Signature}
\label{sec3}
In this section, one novel signature is introduced with two components: patch-based features extracted from the UR2D system; the proposed soft facial attributes extracted with one adapted state-of-the-art CNN to learn 40 facial attributes.

%

\subsection{Patch-based feature component: $\mathbb{S}^{P}$}
To integrated with the UR2D system \cite{xiang2017ijcb}, the same pipeline is followed for pre-processing and extracting the patch-based feature component: Pose Robust Face Signature (PRFS) component \cite{dou2015pose} and Deep Pose Robust Face Signature (DPRFS) component.

Given an input face image, the pipeline of UR2D follows: face detection, landmark detection, pose estimation, 3D reconstruction, texture lifting, and signature extraction. Here only the signature extraction part is introduced, please refer to \cite{xiang2017ijcb} for more details. Both PRFS and DPRFS are extracted from texture lifted images. Facial texture lifting is a technique \cite{Kakadiaris2017137} that lifts the pixel values from the original 2D images to a UV map. Given an original image $I$, a 3D-2D projection matrix $J$, 3D AFM model $M$, it first generates the geometry image $G$, each pixel of which captures the information of an existing or interpolated vertex on the 3D AFM surface. With $G$, a set of 2D coordinates referring to the pixels on an original 2D facial image is computed. Thus, the facial appearance is lifted and represented into a new texture image $T$. A 3D model $M$ and Z-Buffer technique are applied to estimate the occlusion status for each pixel. This process also generates an occlusion mask $M$.

Both PRFS and DPRFS are patch-based features. In PRFS, the facial texture $T$ and the self-occlusion mask $Z$ are first divided into 64 non-overlapping local patches. Then, on each local patch, the discriminative DFD features \cite{lei2014learning} are extracted. In DPRFS, the facial texture $T$ and the self-occlusion mask $Z$ are first divided into eight partially-overlapping local patches. Then, a DPRFS model is trained for each patch based on softmax loss and center loss. Each patch-based feature component contains two part: feature matrix and occlusion encoding. Let $F = \{ f_{ij} \}^{n\times m}$ represent a feature matrix, where each value $f_{ij}$ represent the $i^{th}$ feature of the $j^{th}$ patch while $n$ and $m$ represent the number of features and the number of patches, respectively. The occlusion encoding is represented by $O = \{o_1, o_2, ...,o_m\}$, where $o_j$ is a binary value indicating whether the $j^{th}$ patch is non-occluded. Based on the occlusion encoding of each patch, all the features are combined selectively during matching. Let $\mathbb{S}^{P}=\{F, O\}$ represent the patch-based feature component based on texture-lifted image. The $\mathbb{S}^{P}$ size for PRFS and DPRFS are $64 \times 1024 + 64$, $8 \times 512 + 8$, respectively. Benefit from on CNN features, DPRFS performs better than PRFS.

\subsection{Soft facial attribute component: $\mathbb{S}^{A}$}

\begin{figure} 
	\centering
	\begin{center}
		\includegraphics[width=0.3\linewidth]{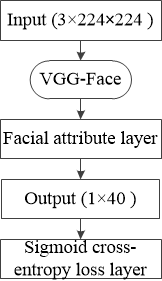}
	\end{center}
	\caption{Depicted is the VGG-Face based network architecture of the soft facial attribute component. }
	\label{f2}
\end{figure}

Given the original 2D image $I$, a CNN is built to extract facial attributes. One state-of-the-art network, such as VGG-Face \cite{parkhi2015deep} or ResNet \cite{masi16dowe} is adapted to learn the facial attribute features. First, the last fully connected layer is removed from the network. Then, a new fully connected facial attribute layer is added which outputs 40 facial attributes, listed in Table~\ref{att_list}. Then, the sigmoid cross-entropy loss is applied to compute the attribute loss over $N$ training images. The network architecture of the proposed facial attribute signature component is depicted in Figure~\ref{f2} with VGG-Face as example. Let $A=\{a_1,a_2, ..., a_{40} \}$ represent the output of facial attribute layer for image $I$. With the sigmoid function, the probability of each facial attribute is obtained, which is denoted as $P=\{p_1,p_2, ..., p_{40} \}$, so that:
    \begin{equation}\label{sm1}
    p_i = \frac{1}{1 + e^{-a_i}}.  \\
    \end{equation} 
Thus, the explicit feature value for each facial attribute is obtained. For each 2D face image, the probability of each facial attribute is computed. This provides us the confidence score of each attribute. By setting a threshold 0.5 on $P$, a binary attribute vector $B=\{b_1,b_2, ... b_{40} \}$ is obtained, that indicates the valid attributes of each facial image. During the matching, the facial attributes that are contributing to improve the performance of face recognition can be directly observed as in Figure~\ref{f1}. Let $\mathbb{S}^{A}=\{A, B\}$ represent the facial attribute signature component. The component size of $\mathbb{S}^{A}$ is $40 \times 2$. The proposed new signature is represented by $\mathbb{S} = \{\mathbb{S}^{D}, \mathbb{S}^{A}\}$. The procedure of signature generation for $\mathbb{S}$ is summarized in Algorithm \ref{a1}.

\begin{table}
	\begin{center}
	\caption{The 40 soft facial attributes are depicted below.}
	\begin{tabular}{ l l} 
		\hline 
	5 O'Clock Shadow  	&Male \\
	Arched Eyebrows  	&Mouth Slightly Open \\
	Attractive 	&Mustache \\
	Bags Under Eyes 	&Narrow Eyes \\
	Bald 	&No Beard \\
	Bangs 	&Oval Face \\
	Big Lips 	&Pale Skin \\
	Big Nose 	&Pointy Nose \\
	Black Hair 	&Receding Hairline \\
	Blond Hair 	&Rosy Cheeks \\
	Blurry 	&Sideburns \\
	Brown Hair 	&Smiling \\
	Bushy Eyebrows 	&Straight Hair\\
	Chubby 	&Wavy Hair \\
	Double Chin 	&Wearing Earrings \\
	Eyeglasses 	&Wearing Hat \\
	Goatee 	&Wearing Lipstick \\
	Gray Hair 	&Wearing Necklace \\
	Heavy Makeup 	&Wearing Necktie\\
	High Cheekbones 	& Young \\
	\hline 
	\end{tabular}
	\label{att_list}
	\end{center}
\end{table}

%

\begin{algorithm}
	\caption{Signature: $\mathbb{S} = \{\mathbb{S}^{P}, \mathbb{S}^{A}\}$}
	\label{a1}
	\KwIn{ 2D image $I$ and 3D AFM model $M$}
	\KwOut{ $\mathbb{S} = \{\mathbb{S}^{P}, \mathbb{S}^{A}\}$}
	Face detection and landmark detection \\
	Pose estimation and 3D reconstruction to obtain $J$ \\	
	Generate geometry image $G$ \\
	Compute texture lifted image $T$ and occlusion mask $M$\\
	Compute feature matrix $F$ and occlusion encoding $O$ \\
	Compute facial attribute information $A$ and $B$ \\
	\Return{\{$\mathbb{S} = \{\mathbb{S}^{P}, \mathbb{S}^{A}\}$, $\mathbb{S}^{P}=\{F, O\}$, $\mathbb{S}^{A}=\{A, B\}$\}};
\end{algorithm}

\section{Signature Matching}
\label{sec4}
\subsection{Patch-based feature component matching}
In this subsection, the matching score is computed based on the component of patch-based features. Following the UR2D system, the cosine score is used to measure the similarity between each pair of signature components computed from gallery and probe image. 

Let $I^g$ and $I^p$ represent a pair of gallery and probe image. In their patch-based signature component $\mathbb{S}^{Pg}$ and $\mathbb{S}^{Pp}$, the feature matrix and occlusion encoding are represented by $F^g$, $F^p$ and $O^g$, $O^p$, respectively. Let $s^p$ represent the score of the patch-based feature component. The features are patch-based, and only non-occluded patches contribute to $s^p$. The signature matching score of patch-based feature component $s^p$ is computed as:
\begin{equation}\label{fsg}
s^p=\frac{1}{k}\sum\limits_{j=1}^{m}(o^g_j \& o^p_j)\times cosine(F^g_j, F^p_j),  \\
\end{equation} 
where $k$ represents the number of non-occluded patch pairs. 

\subsection{Soft facial attribute component matching}

In the facial attribute component, the attribute vectors of $I^g$ and $I^p$ are represented by $A^g$ and $A^p$, respectively. Let $s^a$ represent the score of the facial attribute component, which can be computed directly from
\begin{equation}\label{fsa}
s^a=cosine(A^g, A^p).  \\
\end{equation} 

The final matching score $s$ is computed as the sum of the scores of both components,
\begin{equation}\label{fs}
s = s^p + \lambda s^a,  \\
\end{equation} 
where $\lambda$ represents the weight of the facial attribute component. With this matcher, the contributions are combined of both patch-based features and facial attribute features. $\lambda$ is used to control the contribution of facial attribute features. 

The problem of the previous matching is that all the facial attributes are treated equally. The difference between them are overlooked. However, these different attributes may have different weights. For example, the weight of ``Bags under eyes'' should be larger than that of ``Eyeglasses''. Also, the weight of ``Receding hairline'' should be larger than that of ``Black hair'' or `Blond hair''. Let $W = \{w_1, w_2, ...,w_{40}\}$ represent the weight vector of each attribute. These weights are introduced using the weighted cosine similarity. For $A^g$ and $A^p$, the weighted similarity is computed as:
\begin{equation}\label{sm5}
cosine_w(A^g, A^p, W)  = \frac{\sum\limits_{i=1}^{n}w_ia_i^ga_i^p}{\sqrt{\sum\limits_{i=1}^{n}w_i{a_i^g}^2}\sqrt{\sum\limits_{i=1}^{n}w_i{a^p_i}^2}}.  \\
\end{equation} 
Thus, the weighted attribute matching score can be computed as:
\begin{equation}\label{fsa1}
s^a_w=cosine_w(A^g, A^p, W).  \\
\end{equation} 
The final signature matching score with weighted attribute is:
\begin{equation}\label{fs1}
s_w = s^p + \lambda s^a_w.  \\
\end{equation} 

With this weighted attribute matcher, different weights can be applied to different facial attributes. Note that, if binary weights are applied to facial attributes, the process has the effect of attribute selection. Only the attributes with non-zero weights will be selected in signature matching. The procedure of the signature matching is summarized in Algorithm \ref{a2}.
%
%

\begin{algorithm}
	\caption{Signature matching}
	\label{a2}
	\KwIn{Gallery image signature $\mathbb{S}^{g} = \{\mathbb{S}^{Pg}:\{F^g, O^g\}, \mathbb{S}^{Ag}:\{A^g, B^g\}\}$, probe image signature $\mathbb{S}^{p} = \{\mathbb{S}^{Pp}:\{F^p, O^p\}, \mathbb{S}^{Ap}:\{A^p, B^p\}\}$, $\lambda$ and $W$}
	\KwOut{final siganture matching score $s_w$}
	Compute signature component matching score $s^p$ based on Eq.(\ref{fsg}) \\
	Compute signature component matching score $s^a_w$ based on Eq.(\ref{fsa1}) \\ 
	Compute final matching score $s_w$ based on Eq.(\ref{fs1})\\
	\Return{\{$s_w$\}};
\end{algorithm}


\section{Experiments}
\label{sec5}
This section presents the evaluation of the proposed signature and matcher on two types of face recognition scenarios: constrained environment and unconstrained environment. The datasets used for testing are the UHDB31 dataset \cite{ha2017uhdb31} and the IJB-A dataset \cite{klare2015pushing}. The latest UR2D is used as a baseline pipeline. Following Xu \textit{et al.} \cite{xiang2017ijcb}, the results are also compared with VGG-Face, FaceNet, and COTS v1.9. To demonstrate that the proposed signature can work with different facial features, two different facial features are used: PRFS and DPRFS. The facial attribute networks are trained on the CelebA dataset \cite{liu2015faceattributes}. The weights of pre-trained models are used to fine-tune VGG-Face and ResNet. Both networks are trained for 50,000 iterations with Caffe \cite{jia2014caffe}. The proposed signature with facial attribute is represented as UR2D-A. The $\lambda$ for the signature matcher is set to 0.1, which is learned from a third dataset CASIA WebFace in the range of $\{0.1, 0.2, ..., 1\}$. The weight vector of the Weighted attribute matcher (UR2D-A-W) is decided by the training accuracy of each attribute. A baseline matcher is also created where different probe images are assigned with different weights for different attributes. The weight vector of the weighted Probe attribute matcher (UR2D-A-P) is decided by the attribute confidence scores of each probe image. Rank-1 accuracy is used as performance measurement.  

\subsection{Constrained face recognition}
\begin{figure} 
     \centering
   \begin{center}
     \includegraphics[width=1\linewidth]{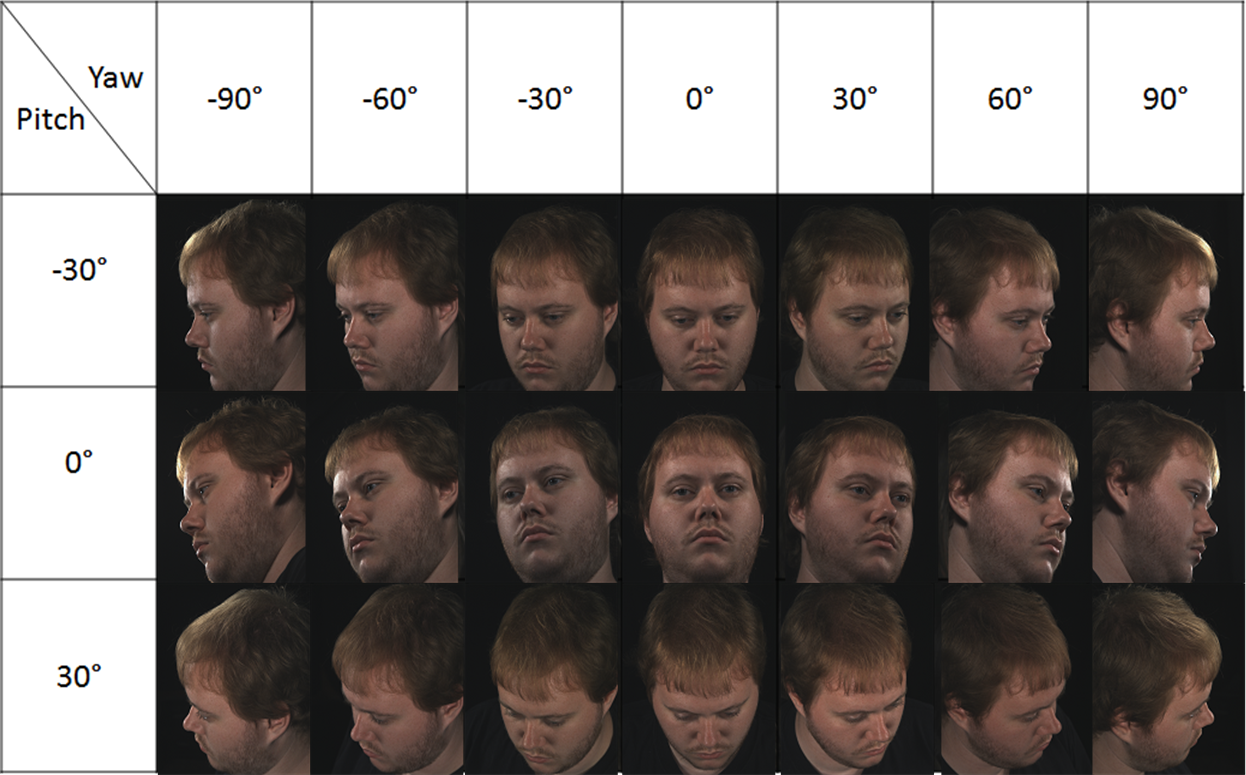}
   \end{center}
      \caption{Depicted are image examples of different poses in the UHDB31 dataset.}
   \label{uhdb31_ex}
   \end{figure}

The UHDB31 dataset \cite{ha2017uhdb31,zhang2017ijcb} contains 29,106 color face images of 77 subjects with 21 poses and 18 illuminations. To exclude the illumination changes, a subset with nature illumination is selected. To evaluate the performance of cross pose face recognition, the front pose (pose-11) face images are used as gallery and the remaining images from 20 poses are used as probe. Figure~\ref{uhdb31_ex} shows the example images from different poses. The performance of different methods under different facial features is shown in Tables \ref{uhdb31_r1} and  \ref{uhdb31_r2}.

%
%
%
 \begin{table*}
  		\caption{Rank-1 performance of different methods computed on the UHDB31 dataset (I)(\%). The methods in row-first order are VGG-Face, COTS v1.9, FaceNet, UR2D-PRFS, UR2D-A-PRFS-VGG-Face, UR2D-A-W-PRFS-VGG-Face, UR2D-A-P-PRFS-VGG-Face, UR2D-A-PRFS-ResNet, UR2D-A-W-PRFS-ResNet and UR2D-A-P-PRFS-ResNet.}
 		\vspace{-7px} 
 		\scalebox{0.5}{ 
 	\begin{tabular}{| c|c |c| c |c |c| c| c|} 
 	\hline 
 	\backslashbox{Pitch}{Yaw}
 	& -90\textdegree{} &-60\textdegree{} &-30\textdegree{} &0\textdegree{} & +30\textdegree{} &+60\textdegree{} &+90\textdegree{} \\
 	\hline
 	+30\textdegree{} &
 	\begin{tabular}{c c c } 14 & 11 & {\bf 58} \\  48& 51&48 \\ 49& 52&52 \\ 48 && \end{tabular} &
 	\begin{tabular}{c c c } 69 & 32 & {\bf 95} \\  90& 94&94 \\ 93& 94&94 \\ 94 && \end{tabular} &
 	\begin{tabular}{c c c } 94&90&{\bf 100} \\ {\bf 100}& {\bf 100}&{\bf 100} \\ {\bf 100}&{\bf 100}&{\bf 100} \\ {\bf 100} && \end{tabular} &
 	\begin{tabular}{c c c } 99&{\bf 100}&{\bf 100}\\{\bf 100}&{\bf 100}&{\bf 100}\\{\bf 100}&{\bf 100}&{\bf 100}\\ {\bf 100} && \end{tabular} &
 	\begin{tabular}{c c c } 95&93&99 \\ {\bf 100}&{\bf 100}&{\bf 100} \\ {\bf 100}&{\bf 100}&{\bf 100}\\{\bf 100} && \end{tabular} &
 	\begin{tabular}{c c c } 79&38&92 \\ 95& 96&96 \\ {\bf 97}& 96&96 \\ 96 && \end{tabular} &
 	\begin{tabular}{c c c } 19&7&{\bf 60} \\ 47&55&51 \\ 52&51&49 \\ 51 && \end{tabular} \\
 		\hline
 	0\textdegree{} &
 	\begin{tabular}{c c c } 22&9&{\bf 84}\\ 79&82&82 \\ 82&{\bf 84}&{\bf 84} \\ 82 && \end{tabular} &
 	\begin{tabular}{c c c } 88&52&99\\{\bf 100}&{\bf 100}&{\bf 100}\\{\bf 100}&{\bf 100}&{\bf 100}\\{\bf 100} && \end{tabular} &
 	\begin{tabular}{c c c } {\bf 100}&99&{\bf 100}\\{\bf 100}&{\bf 100}&{\bf 100}\\{\bf 100}&{\bf 100}&{\bf 100}\\{\bf 100} && \end{tabular} &
 	- &
 	\begin{tabular}{c c c } {\bf 100}&{\bf 100}&{\bf 100}\\{\bf 100}&{\bf 100}&{\bf 100}\\{\bf 100}&{\bf 100}&{\bf 100}\\{\bf 100} && \end{tabular} &
 	\begin{tabular}{c c c } 94&73&99\\{\bf 100}&{\bf 100}&{\bf 100}\\{\bf 100}&{\bf 100}&{\bf 100}\\{\bf 100} && \end{tabular} &
 	\begin{tabular}{c c c } 27&10&91\\84&87&87\\86&{\bf 88}&87\\87 && \end{tabular} \\
	\hline	
 	-30\textdegree{} &
 	\begin{tabular}{c c c } 8&0&44\\43&45&46\\{\bf 47}&44&{\bf 47}\\46 && \end{tabular} &
 	\begin{tabular}{c c c } 2&19&80\\90&{\bf 92}&{\bf 92}\\{\bf 92}&{\bf 92}&{\bf 92}\\{\bf 92} && \end{tabular} &
 	\begin{tabular}{c c c } 91&90&{\bf 99}\\{\bf 99}&{\bf 99}&{\bf 99}\\{\bf 99}&{\bf 99}&{\bf 99}\\{\bf 99} && \end{tabular} &
 	\begin{tabular}{c c c } 96&99&99\\{\bf 100}&{\bf 100}&{\bf 100}\\{\bf 100}&{\bf 100}&{\bf 100}\\{\bf 100} && \end{tabular} &
 	\begin{tabular}{c c c } 96&98&97\\{\bf 99}&{\bf 99}&{\bf 99}\\{\bf 99}&{\bf 99}&{\bf 99}\\{\bf 99}&& \end{tabular} &
 	\begin{tabular}{c c c } 52&15&90\\95&{\bf 97}&{\bf 97}\\{\bf 97}&{\bf 97}&{\bf 97}\\{\bf 97} && \end{tabular} &
 	\begin{tabular}{c c c } 9&3&35\\{\bf 58}&57&57\\57&{\bf 58}&{\bf 58}\\57 && \end{tabular} \\
 			\hline
 \end{tabular}}
 	\label{uhdb31_r1}
  \end{table*}

\begin{table*}
 
		\caption{Rank-1 performance of different methods computed on the UHDB31 dataset (II)(\%). The methods in row-first order are VGG-Face, COTS v1.9, FaceNet, UR2D-DPRFS, UR2D-A-DPRFS-VGG-Face, UR2D-A-W-DPRFS-VGG-Face, UR2D-A-P-DPRFS-VGG-Face, UR2D-A-DPRFS-ResNet, UR2D-A-W-DPRFS-ResNet and UR2D-A-P-DPRFS-ResNet.}
		\vspace{-7px} 
		\scalebox{0.5}{ 
	\begin{tabular}{| c|c |c| c |c |c| c| c|} 
	 	\hline 
	 	\backslashbox{Pitch}{Yaw}
	 	& -90\textdegree{} &-60\textdegree{} &-30\textdegree{} &0\textdegree{} & +30\textdegree{} &+60\textdegree{} &+90\textdegree{} \\
	 	\hline
	 	+30\textdegree{} &
	 	\begin{tabular}{c c c } 14&11&58\\82&82&{\bf 83}\\82&82&{\bf 83}\\{\bf 83} && \end{tabular} &
	 	\begin{tabular}{c c c } 69&32&95\\{\bf 99}&{\bf 99}&{\bf 99}\\{\bf 99}&{\bf 99}&{\bf 99}\\{\bf 99} && \end{tabular} &
	 	\begin{tabular}{c c c } 94&90&{\bf 100}\\{\bf 100}&{\bf 100}&{\bf 100}\\{\bf 100}&{\bf 100}&{\bf 100}\\{\bf 100} && \end{tabular} &
	 	\begin{tabular}{c c c } 99&{\bf 100}&{\bf 100}\\{\bf 100}&{\bf 100}&{\bf 100}\\{\bf 100}&{\bf 100}&{\bf 100}\\{\bf 100} && \end{tabular} &
	 	\begin{tabular}{c c c } 95&93&{\bf 99}\\{\bf 99}&{\bf 99}&{\bf 99}\\{\bf 99}&{\bf 99}&{\bf 99}\\{\bf 99} && \end{tabular} &
	 	\begin{tabular}{c c c } 79&38&92\\{\bf 99}&97&{\bf 99}\\97&97&{\bf 99}\\{\bf 99} && \end{tabular} &
	 	\begin{tabular}{c c c } 19&7&60\\75&74&75\\75&{\bf 78}&77\\77 && \end{tabular} \\
	 		\hline
	 	0\textdegree{} &
	 	\begin{tabular}{c c c } 22&9&84\\96&{\bf 97}&{\bf 97}\\{\bf 97}&{\bf 97}&96\\96 && \end{tabular} &
	 	\begin{tabular}{c c c } 88&52&99\\{\bf 100}&{\bf 100}&{\bf 100}\\{\bf 100}&{\bf 100}&{\bf 100}\\{\bf 100} && \end{tabular} &
	 	\begin{tabular}{c c c } {\bf 100}&99&{\bf 100}\\{\bf 100}&{\bf 100}&{\bf 100}\\{\bf 100}&{\bf 100}&{\bf 100}\\{\bf 100} && \end{tabular} &
	 	- &
	 	\begin{tabular}{c c c } {\bf 100}&{\bf 100}&{\bf 100}\\{\bf 100}&{\bf 100}&{\bf 100}\\{\bf 100}&{\bf 100}&{\bf 100}\\{\bf 100} && \end{tabular} &
	 	\begin{tabular}{c c c } 94&73&99\\{\bf 100}&{\bf 100}&{\bf 100}\\{\bf 100}&{\bf 100}&{\bf 100}\\{\bf 100} && \end{tabular} &
	 	\begin{tabular}{c c c } 27&10&91\\{\bf 96}&{\bf 96}&{\bf 96}\\{\bf 96}&{\bf 96}&{\bf 96}\\{\bf 96} && \end{tabular} \\
		\hline	
	 	-30\textdegree{} &
	 	\begin{tabular}{c c c } 8&0&44\\75&{\bf 78}& 77\\76&76&76\\76 && \end{tabular} &
	 	\begin{tabular}{c c c } 2&19&80\\97&{\bf 99}&{\bf 99}\\{\bf 99}&{\bf 99}&{\bf 99}\\{\bf 99} && \end{tabular} &
	 	\begin{tabular}{c c c } 91&90&99\\{\bf 100}&{\bf 100}&{\bf 100}\\{\bf 100}&{\bf 100}&{\bf 100}\\{\bf 100} && \end{tabular} &
	 	\begin{tabular}{c c c } 96&99&99\\{\bf 100}&{\bf 100}&{\bf 100}\\{\bf 100}&{\bf 100}&{\bf 100}\\{\bf 100} && \end{tabular} &
	 	\begin{tabular}{c c c } 96&98&97\\{\bf 100}&{\bf 100}&{\bf 100}\\{\bf 100}&{\bf 100}&{\bf 100}\\{\bf 100}&& \end{tabular} &
	 	\begin{tabular}{c c c } 52&15&90\\{\bf 96}&{\bf 96}&{\bf 96}\\{\bf 96}&{\bf 96}&95\\95 && \end{tabular} &
	 	\begin{tabular}{c c c } 9&3&35\\79&{\bf 83}&{\bf 83}\\{\bf 83}&82&{\bf 83}\\{\bf 83} && \end{tabular} \\
	 			\hline

 \end{tabular}}
 		\label{uhdb31_r2}
\end{table*}

From Table \ref{uhdb31_r1} and \ref{uhdb31_r2}, it can be observed that with PRFS feature, the proposed signature can improve accuracy under nine poses, especially some large poses like pose-1 to pose-4 and pose-18 to pose-21. The accuracy improvements range from 1\% to 8\%. At the same time, the excellent performance of close frontal poses is retained. Under DPRFS, the proposed signature also achieves better performance. Overall, the proposed signatures achieve the best results on most of poses. The accuracy improvements range from 1\% to 4\%. It can be observed that the performance of facial attributes based on VGG-Face and ResNet are comparable. In addition, the weighed attribute matching and the weighted probe matching achieve comparable results on all the poses.

\subsection{Unconstrained face recognition}
The IJB-A dataset \cite{klare2015pushing} contains images and videos from 500 subjects captured from ``in the wild'' environment. This dataset merges images and frames and provides evaluations on the template level. A template contains one or several images/frames of one subject. According to the IJB-A protocol, it splits galleries and probes into 10 splits. In the experiment, the same modification as \cite{xiang2017ijcb} is followed for use it in close-set face recognition. The performance of different methods under different global signatures is shown in Table \ref{ijba_r1}.

\begin{table*}
	\begin{center}
	\caption{Rank-1 performance of different methods computed on the IJB-A dataset (\%).}
	\vspace{-7px}
   \scalebox{0.6}{
	\begin{tabular}{ l |c c c c c c c c c c c} 
	\hline 
	Methods &split-1 & split-2 &split-3 &split-4 &split-5 & split-6 & split-7 &split-8 & split-9 & split-10 & Average\\
		\hline 
    VGG-Face &76.18 &74.37 &24.33 &47.67 &52.07 &47.11 &58.31 &54.31 &47.98 &49.06 &53.16 \\
    COTS v1.9 &75.68 &76.57 &73.66 &76.73 &76.31 &77.21 &76.27 &74.50 &72.52 &77.88 &75.73\\
    	\hline 
    UR2D-PRFS &49.01	&49.57	&48.22	&47.75	&48.85	&44.46	&52.46	&48.22	&43.48	&48.79	&48.08\\
        \hline 
    UR2D-A-PRFS-VGG-Face &52.77&52.83	&{\bf 51.16}	&  51.08 	&51.21	&47.42	& 56.47 	&  51.26 	&{\bf 46.24}	&52.23	&51.27\\

    UR2D-A-W-PRFS-VGG-Face &52.85&52.80	&{\bf 51.16}	&{\bf 51.12}	&{\bf 51.42}	&47.38	&{\bf 56.78}	&{\bf 51.50}	& 46.11 	&52.13	&51.33\\
        UR2D-A-P-PRFS-VGG-Face &52.13&52.27	& 50.46	& 50.14 	& 50.79 	&46.70	& 55.62	& 50.86	& 45.41 	&51.23	&50.56\\
        \hline 
    UR2D-A-PRFS-ResNet& 52.95	&{\bf 53.57}	&50.97	&51.05	&51.22	&  47.75 	&56.38	&51.16	&46.22	&{\bf 52.41}	&  51.37 \\
    UR2D-A-W-PRFS-ResNet&{\bf 53.05}	& 53.50	&  51.05 	&51.08	& 51.32 	&{\bf 47.83}	&56.08	&51.43	&46.07	& 52.33 	&{\bf 51.38}\\
        UR2D-A-P-PRFS-ResNet&  51.95 	& 52.59	&  50.66 	&50.33	& 50.65 	&  46.82	&55.38	&50.80	&45.25	& 51.16 	& 50.56\\
   	\hline     
    UR2D-DPRFS &78.78&77.60&	77.94&	79.88&	78.44&	80.57&	81.78&	79.00&	75.94&	79.22&	78.92 \\
    \hline 
    UR2D-A-DPRFS-VGG-Face &  79.16 	&77.80	&  78.38 	&80.04	&78.51	&80.81&	81.96&	79.14&	76.16&	79.36&	79.13\\
    UR2D-A-W-DPRFS-VGG-Face &{\bf 79.19}	&77.82	&{\bf 78.53}	&80.04	&	{\bf 78.76}	&	{\bf 80.95}&	81.78&	79.37&		{\bf 76.50}&	79.51&		{\bf 79.29}\\
        UR2D-A-P-DPRFS-VGG-Face & 79.09 	&77.88	&  78.42	&80.17	&78.72	&	{\bf 80.95}&	81.70&	79.10&	76.35&		{\bf 79.54}&	79.19\\
        \hline 
    UR2D-A-DPRFS-ResNet&78.95	&{\bf 77.92}	&78.23	&{\bf 80.50}	& 78.54	& 80.87	&{\bf 82.16}&  79.54&	  76.24 &  79.51 &	79.25 \\
    
    UR2D-A-W-DPRFS-ResNet&79.06	&{\bf 77.92}	&78.34	& 80.44	& 78.62 	& 80.91 	&82.03&{\bf 	79.61}&	  76.24 &	79.47& 79.26  \\
        UR2D-A-P-DPRFS-ResNet&78.89	& 77.88	&	{\bf 78.53}	& 80.17	& 78.62 	&  80.74	&82.12& 	79.41 &	  76.24 &	79.36&	 79.20 \\
	\hline 
	\end{tabular}}
	\label{ijba_r1}
	\end{center}
\end{table*}

\begin{figure}
 \centering
\begin{subfigure}[b]{0.45\textwidth}
      \includegraphics[width=\textwidth]{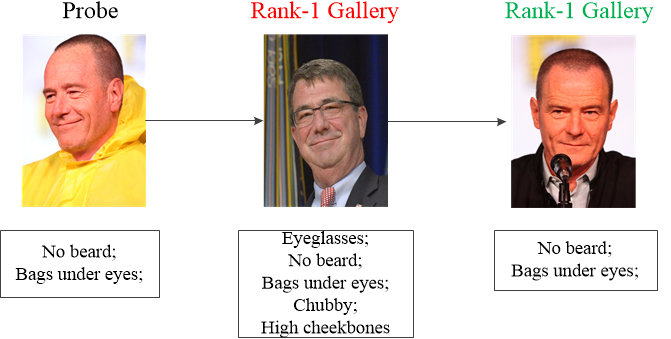}
      \caption{}
\end{subfigure} 
\hfill\hfill%
\begin{subfigure}[b]{0.45\textwidth}
      \includegraphics[width=\textwidth]{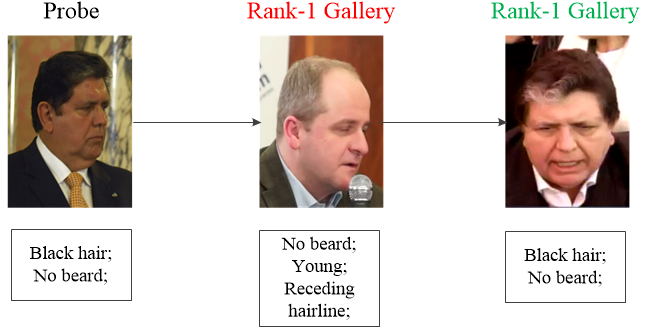}
      \caption{}
\end{subfigure}
  
\begin{subfigure}[b]{0.45\textwidth}
        \includegraphics[width=\textwidth]{fa_ex8}
        \caption{}
\end{subfigure}
\hfill\hfill%
\begin{subfigure}[b]{0.45\textwidth}
\includegraphics[width=\textwidth]{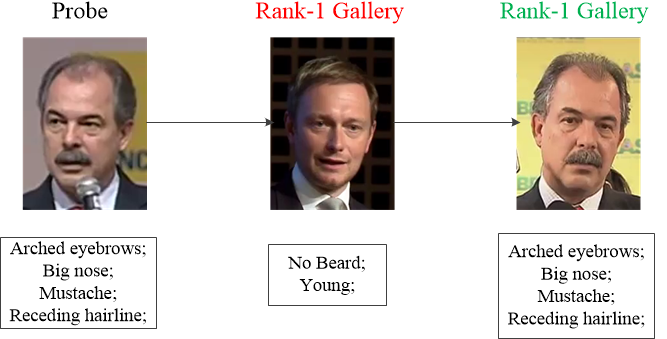}
\caption{}
\end{subfigure}

\begin{subfigure}[b]{0.45\textwidth}
	\includegraphics[width=\textwidth]{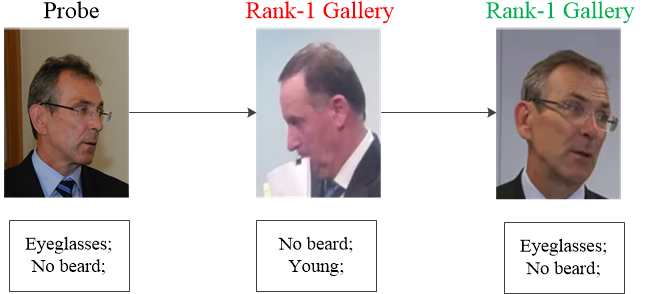}
	\caption{}
\end{subfigure}
                            
   \caption{ Depicted are the matching examples of the proposed signature. The three images in each example represent probe, incorrect gallery image matched with patch-based features only, and correct gallery image matched with the proposed signature, respectively.}\label{f3-ex}
\end{figure}

From Table \ref{ijba_r1}, it can be observed that with PRFS signature, the proposed UR2D-A signature can improve the accuracy under all the splits. The average accuracy is improved by 3.19\% and 3.29\% with UR2D-A-VGG-Face and UR2D-A-ResNet, respectively. Under DPRFS, the proposed UR2D-A signature also achieves better performance. The average accuracy is improved by 0.21\% and 0.33\% with VGG-Face and ResNet, respectively. In addition, the weighed attribute matcher and the weighted probe attribute matcher achieve comparable results on all the splits. Overall, the proposed signature achieves the best results on all the splits compared to previous methods. The best performance is achieved by UR2D-A-W-DPRFS-VGG-Face. Figure \ref{f3-ex} depicts more matching examples of the proposed signature. It can be observed that the proposed signature can be used to correct the matching error of implicit facial features. The reason behind this is that the proposed signature is more robust to facial attribute information. The facial attribute information is well captured to improve the performance while this information is overlooked in previous implicit facial feature based methods.      

\subsection{Sensitivity Analysis}

\begin{figure}
 \centering
  \begin{subfigure}[b]{0.49\textwidth}
          \includegraphics[width=\textwidth]{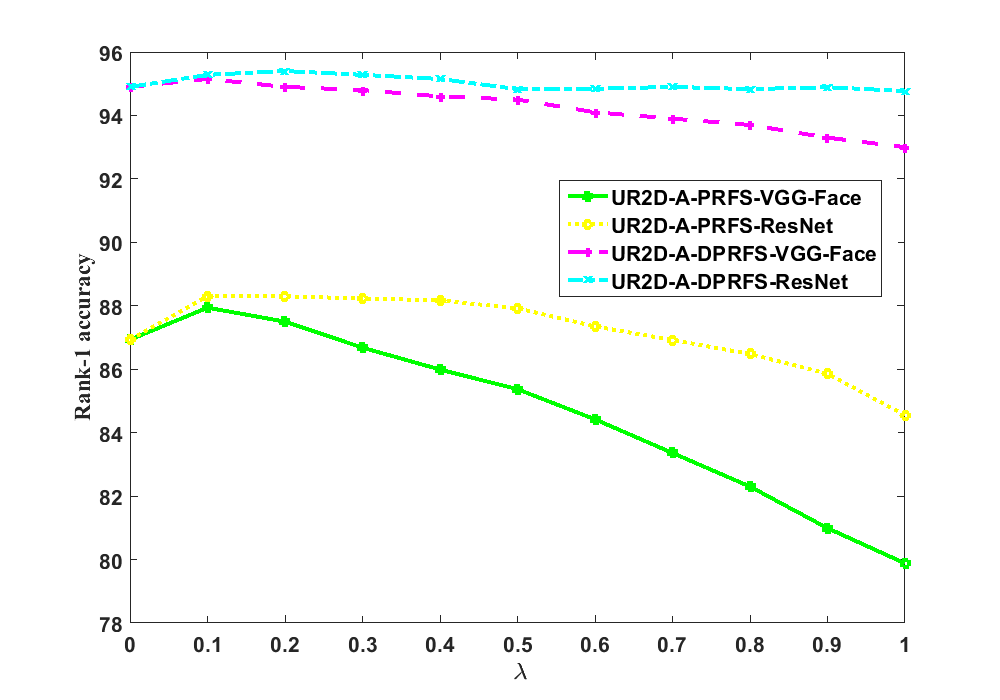}
          \caption{}
  \end{subfigure}
  \hfill%
  \begin{subfigure}[b]{0.49\textwidth}
          \includegraphics[width=\textwidth]{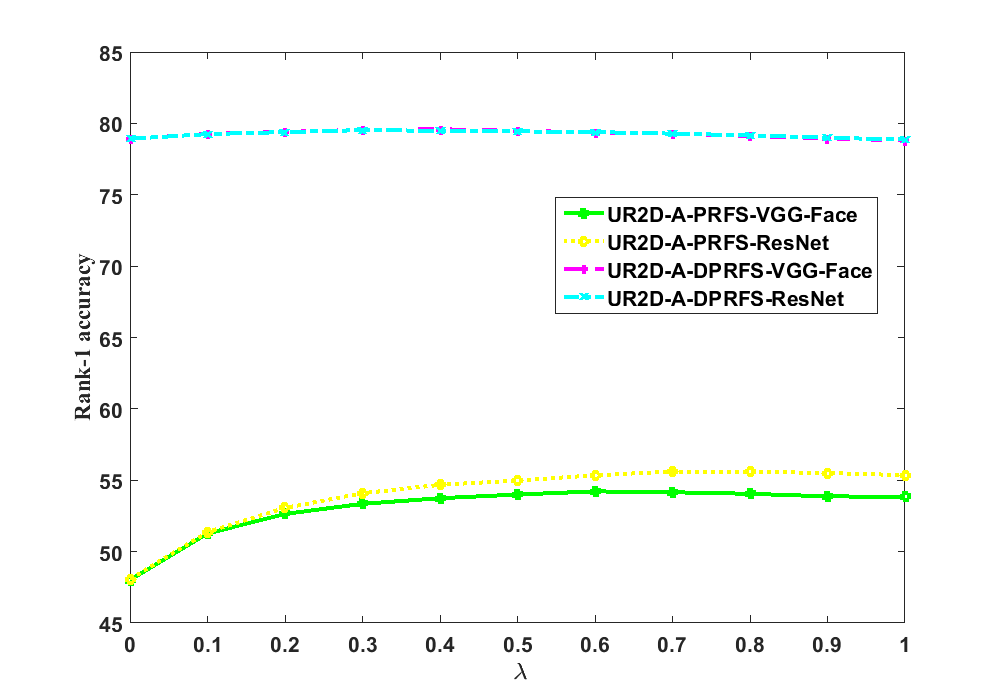}
          \caption{}
  \end{subfigure}

   \caption{ The performance of UR2D-A computed with different $\lambda$ values. (a) UHDB31. (b) IJB-A.}\label{re-f4}
\end{figure}

\begin{figure}
 \centering
  \begin{subfigure}[b]{0.49\textwidth}
          \includegraphics[width=\textwidth]{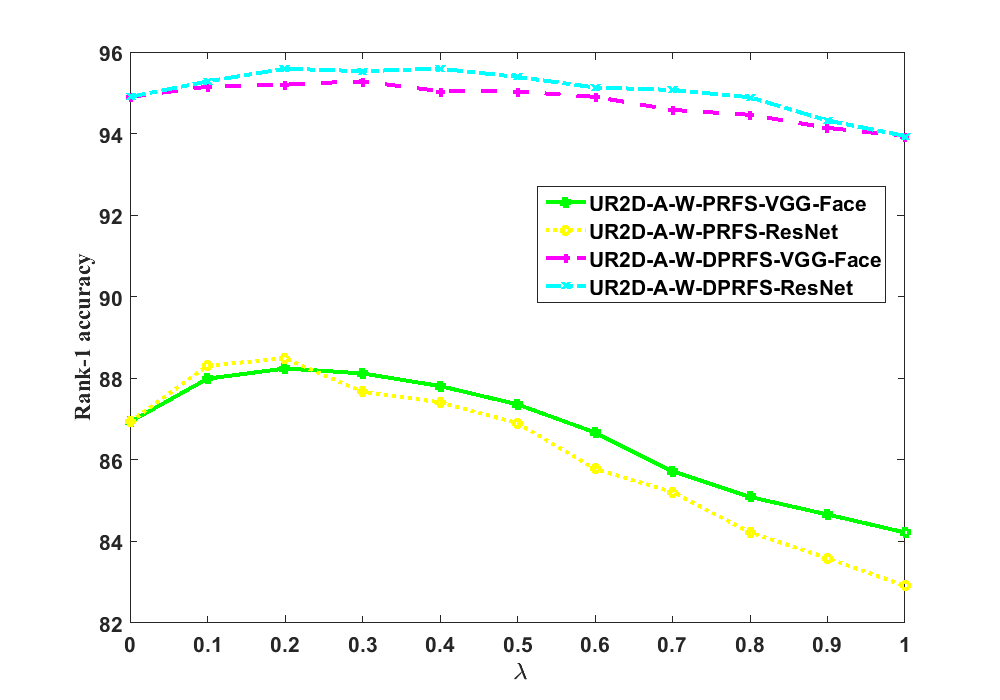}
          \caption{}
  \end{subfigure} 
  \begin{subfigure}[b]{0.49\textwidth}
          \includegraphics[width=\textwidth]{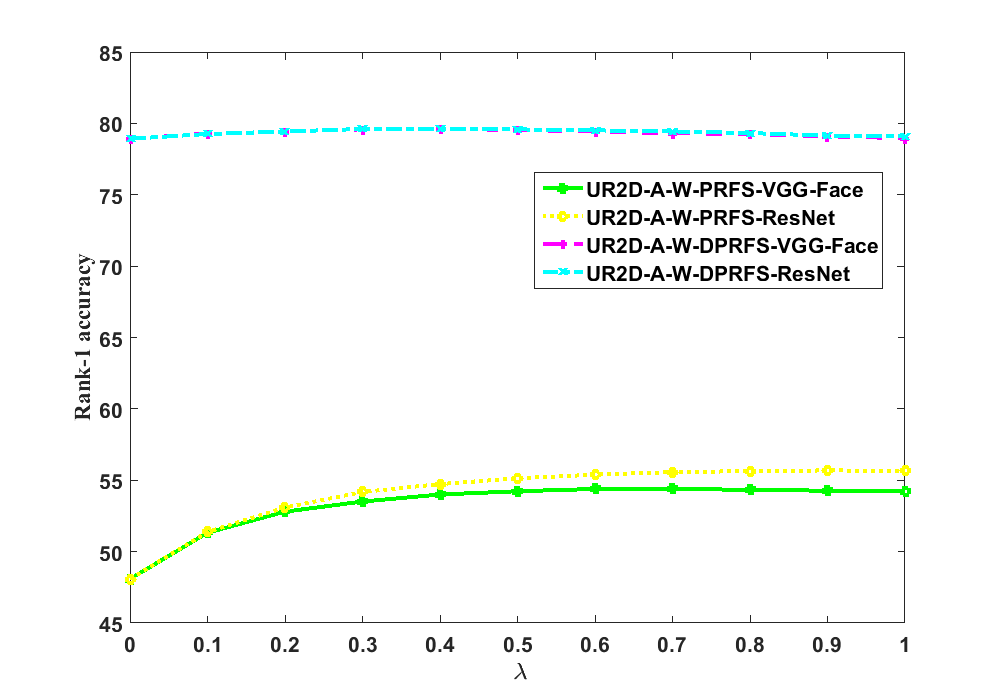}
          \caption{}
  \end{subfigure}

   \caption{ The performance of UR2D-A-W computed with different $\lambda$ values. (a) UHDB31. (b) IJB-A.}\label{re-f5}
\end{figure}

\begin{figure}
 \centering
  \begin{subfigure}[b]{0.49\textwidth}
          \includegraphics[width=\textwidth]{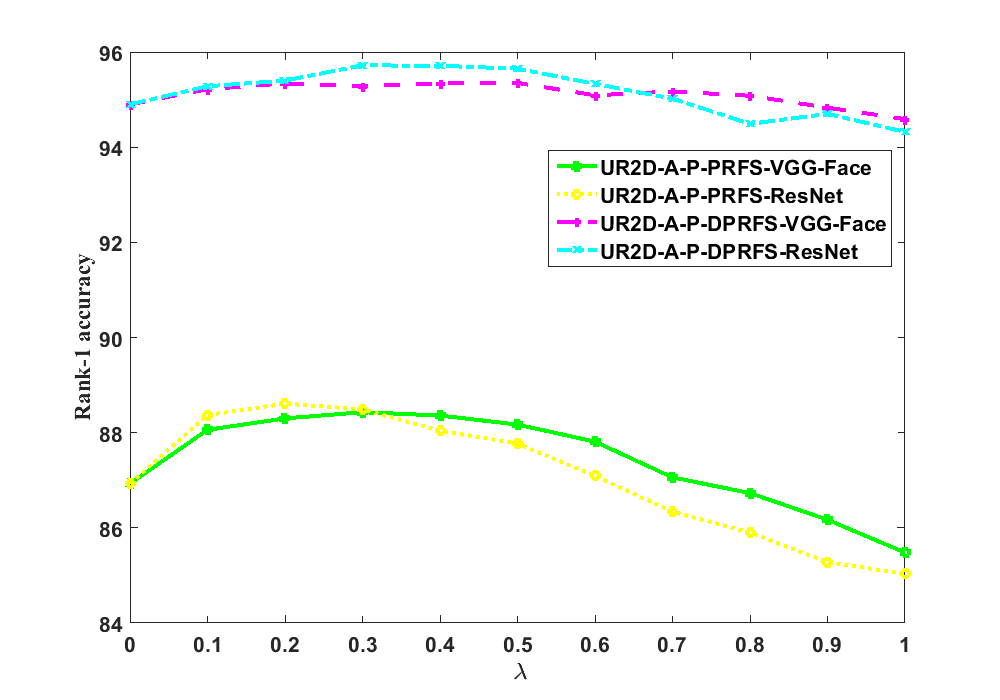}
          \caption{}
  \end{subfigure} 
  \begin{subfigure}[b]{0.49\textwidth}
          \includegraphics[width=\textwidth]{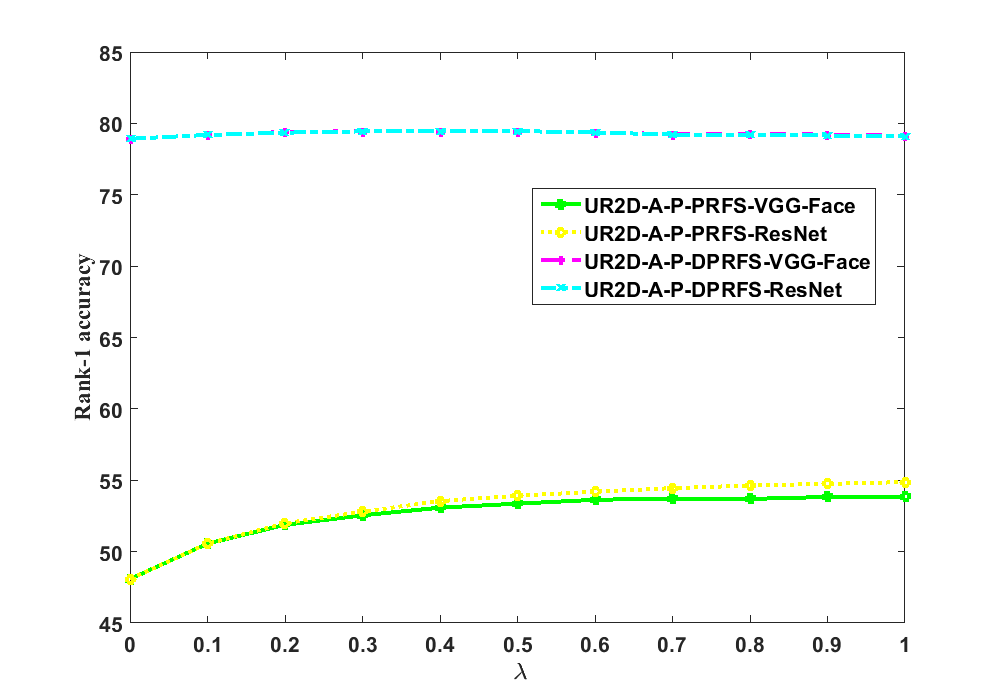}
          \caption{}
  \end{subfigure}

   \caption{ The performance of UR2D-A-P computed with different $\lambda$ values. (a) UHDB31. (b) IJB-A.}\label{re-f6}
\end{figure}

In this section, the sensitivity of $\lambda$ is evaluated with different values in the range of $\{0.1, 0.2, ..., 1\}$ for different matchers. The results of UR2D-A, UR2D-A-W and UR2D-A-P are shown in Figures~\ref{re-f4}-\ref{re-f6}, respectively.

It can be observed that UR2D-A, UR2D-A-W and UR2D-A-P perform similarly on the two datasets. Different methods achieve the best result with different $\lambda$ values. Also, the performance of DPRFS is less sensitive to $\lambda$ than that of PRFS.

\subsection{Statistical Analysis}
\label{SA}    

In this section, statistical analysis is performed for the baseline and the best version of the proposed signature and matcher (UR2D-DPRFS and UR2D-A-W-DPRFS-VGG-Face) over the 30 data splits (20 from UHDB31 and 10 from IJB-A). From Dem\v{s}ar \textit{et al.} \cite{demvsar2006statistical}, the Friedman test \cite{friedman1937use,friedman1940comparison} and the two tailed Bonferroni-Dunn test \cite{dunn1961multiple} are used to compare multiple methods over multiple datasets. Let $r_i^j$ represent the rank of the $j^{th}$ of k algorithm on the $i^{th}$ of $N$ datasets.  The Friedman test compares the average ranks of different methods, by $R_j = \frac{1}{N} \sum_i r_i^j$. The null-hypothesis states that all the methods are equal, so their ranks $R_j$ should be equivalent. The original Friedman statistic \cite{friedman1937use,friedman1940comparison}, 
    \begin{equation}\label{st1}
     \mathcal{X}_F^2 = \frac{12N}{k(k+1)}[\sum_j R_j^2 - \frac{k(k+1)^2}{4}],
    \end{equation}
is distributed according to $\mathcal{X}_F^2$ with $k-1$ degrees of freedom. Due to its undesirable conservative property,  Iman \textit{et al.} \cite{iman1980approximations} derived a better statistic
    \begin{equation}\label{st2}
     F_F = \frac{(N-1)\mathcal{X}_F^2}{N(k-1)-\mathcal{X}_F^2},
    \end{equation}
which is distributed according to the F-distribution with $k-1$ and $(k-1) \times (N-1)$ degrees of freedom. First the average ranks for UR2D-DPRFS and UR2D-A-W-DPRFS-VGG-Face are computed as 1.73 and 1.27, respectively. The $F_F$ statistical values of Rank-1 accuracy based on (\ref{st2}) are computed as $7.78$. With two methods and 30 data splits, $F_F$ is distributed with $2-1$ and $(2-1) \times (30-1) = 29$ degrees of freedom. The critical value of $F(1, 29)$ for $\alpha = 0.10$ is $2.88 < 7.78$, so the null-hypothesis is rejected. Then, the two tailed Bonferroni-Dunn test is applied to compare the two methods by the critical difference:
     \begin{equation}\label{st3}
      CD = q_{\alpha} \sqrt{\frac{k(k+1)}{6N}},
     \end{equation}
where $q_{\alpha}$ is the critical values. If the average rank between two methods is larger than critical difference, the two methods are significantly different. According to Table 5 in \cite{demvsar2006statistical}, the critical value of two methods when $p = 0.10$ is 1.65. The critical difference is computed as $CD = 1.65 \sqrt{\frac{2 \times 3}{6 \times 30}} = 0.30$. In conclusion, under Rank-1 accuracy, UR2D-A-W-DPRFS-VGG-Face performs significantly better than UR2D-DPRFS (the difference between ranks are $1.73 - 1.27 = 0.46 > 0.30$). 
%

\section{Conclusion}
\label{sec6}
This paper proposed a facial signature that contains both implicit facial features and explicit facial attribute features. Explicit soft facial attribute information is extracted to improve the performance of face recognition system that only uses implicit facial features. The experimental results confirmed the assumptions that facial attribute features explore more local discriminative information and can be used to improve matching performance. Comparing with the UR2D system, the Rank-1 accuracy is improved significantly by 4\% and 0.37\% for the UHDB31 dataset and the IJB-A dataset, respectively.

 \section*{Acknowledgements}
 This material is based upon work supported by the U.S. Department of Homeland Security under Grant Award Number 2015-ST-061-BSH001. This grant is awarded to the Borders, Trade, and Immigration (BTI) Institute: A DHS Center of Excellence led by the University of Houston, and includes support for the project ``Image and Video Person Identification in an Operational Environment: Phase I'' awarded to the University of Houston. The views and conclusions contained in this document are those of the authors and should not be interpreted as necessarily representing the official policies, either expressed or implied, of the U.S. Department of Homeland Security.

\section*{References}
\bibliography{egbib_all}

\end{document}